\documentclass{article}

\usepackage{hyperref} 
\usepackage[accepted]{icml2018}

\usepackage{natbib}
\usepackage{graphicx}
\usepackage[latin1]{inputenc}
\usepackage{amssymb,amsmath,amsthm,array,amsfonts}
\usepackage{color}
\usepackage{booktabs}
\usepackage{underoverlap}
\usepackage{subfigure}
\usepackage{multirow} 
\usepackage[rightcaption]{sidecap}
\sidecaptionvpos{figure}{c}

\def\x{{\mathbf x}}
\def\u{{\mathbf u}}
\def\X{{\mathbf X}}
\def\W{{\mathbf W}}
\def\V{{\mathbf V}}
\def\U{{\mathbf U}}
\def\M{{\mathbf M}}
\def\A{{\mathbf A}}
\def\L{{\mathbf L}}
\def\D{{\mathbf D}}
\def\I{{\mathbf I}}
\def\L{{\mathbf L}}

\newtheorem{theorem}{Theorem}

\icmltitlerunning{GNNs with convolutional ARMA filters}

\begin{document}

\twocolumn[
\icmltitle{Graph Neural Networks with Convolutional ARMA Filters}



\icmlsetsymbol{equal}{*}

\begin{icmlauthorlist}
\icmlauthor{Filippo Maria Bianchi}{uit,nor}
\icmlauthor{Daniele Grattarola}{usi}
\icmlauthor{Lorenzo Livi}{man,exe}
\icmlauthor{Cesare Alippi}{usi,poli}
\end{icmlauthorlist}

\icmlaffiliation{uit}{Dept. of Mathematics and Statistics, UiT the Arctic Univeristy of Norway}
\icmlaffiliation{nor}{NORCE, The Norwegian Research Centre}
\icmlaffiliation{usi}{Faculty of Informatics, Universit\`a della Svizzera Italiana, Switzerland}
\icmlaffiliation{poli}{Dept. of Electronics, Information, and Bioengineering, Politecnico di Milano, Italy}
\icmlaffiliation{man}{Dept. of Computer Science and Mathematics, University of Manitoba, Canada}
\icmlaffiliation{exe}{Dept. of Computer Science, University of Exeter, United Kingdom}

\icmlcorrespondingauthor{Filippo Maria Bianchi}{filippombianchi@gmail.com}

\icmlkeywords{Convolutional neural networks, Spectral graph convolution, graph filtering}

\vskip 0.3in
]



\printAffiliationsAndNotice{}  

\begin{abstract}
Popular graph neural networks implement convolution operations on graphs based on polynomial spectral filters. 
In this paper, we propose a novel graph convolutional layer inspired by the auto-regressive moving average (ARMA) filter that, compared to polynomial ones, provides a more flexible frequency response, is more robust to noise, and better captures the global graph structure.
We propose a graph neural network implementation of the ARMA filter with a recursive and distributed formulation, obtaining a convolutional layer that is efficient to train, localized in the node space, and can be transferred to new graphs at test time. 
We perform a spectral analysis to study the filtering effect of the proposed ARMA layer and report experiments on four downstream tasks: semi-supervised node classification, graph signal classification, graph classification, and graph regression.
Results show that the proposed ARMA layer brings significant improvements over graph neural networks based on polynomial filters.
\end{abstract}

\section{Introduction}
Graph Neural Networks (GNNs) are a class of models lying at the intersection between deep learning and methods for structured data, which perform inference on discrete objects (nodes) by accounting for arbitrary relationships (edges) among them~\cite{bronstein2017geometric, battaglia2018relational}.
A GNN combines node features within local neighborhoods on the graph to learn node representations that can be directly mapped into categorical labels or real values~\cite{scarselli2009graph,klicpera2019predict}, or combined to generate graph embeddings for graph classification and regression~\cite{perozzi2014deepwalk, duvenaud2015convolutional, yang2016revisiting, hamilton2017inductive, bacciu2018contextual}.

The focus of this work is on GNNs that implement a convolution in the spectral domain with a non-linear trainable filter~\cite{bruna2013spectral, henaff2015deep}.
Such a filter selectively shrinks or amplifies the Fourier coefficients of the graph signal (an instance of the node features) and then maps the node features to a new space.
To avoid the expensive spectral decomposition and projection in the frequency domain, state-of-the-art GNNs implement graph filters as low-order polynomials that are learned directly in the node domain~\cite{defferrard2016convolutional,kipf2016semi,kipf2016variational}.
Polynomial filters have a finite impulse response and perform a weighted moving average filtering of graph signals on local node neighborhoods~\cite{tremblay2018design}, allowing for fast distributed implementations such as those based on Chebyshev polynomials and Lanczos iterations~\cite{susnjara2015accelerated, defferrard2016convolutional, liao2018lanczos}.
Polynomial filters have limited modeling capabilities~\cite{isufi2016autoregressive} and, due to their smoothness, cannot model sharp changes in the frequency response~\cite{tremblay2018design}.
Crucially, polynomials with high degree are necessary to reach high-order neighborhoods, but they tend to be more computationally expensive and, most importantly, overfit the training data making the model sensitive to changes in the graph signal or the underlying graph structure. 
A more versatile class of filters is the family of Auto-Regressive Moving Average filters (ARMA)~\cite{narang2013signal}, which offer a larger variety of frequency responses and can account for higher-order neighborhoods compared to polynomial filters with the same number of parameters.

In this paper, we address the limitations of existing graph convolutional layers inspired by polynomial filters and propose a novel GNN convolutional layer based on ARMA filters.
Our ARMA layer implements a non-linear and trainable graph filter that generalizes the convolutional layers based on polynomial filters and provides the GNN with enhanced modeling capability, thanks to a flexible design of the filter's frequency response.
The ARMA layer captures global graph structures with fewer parameters, overcoming the limitations of GNNs based on high-order polynomial filters.

ARMA filters are not localized in node space and require to compute a matrix inversion, which is intractable in the context of GNNs.
To address this issue, the proposed ARMA layer relies on a recursive formulation, which leads to a fast and distributed implementation that exploits efficient sparse operations on tensors.
The resulting filters are not learned in the Fourier space induced by a given Laplacian, but are localized in the node space and are independent of the underlying graph structure.
This allows our GNN to handle graphs with unseen topologies during the test phase of inductive inference tasks.

The performance of the proposed ARMA layer is evaluated on semi-supervised node classification, graph signal classification, graph classification, and graph regression tasks. 
Results show that a GNN equipped with ARMA layers outperforms GNNs with polynomial filters in every downstream task.

\section{Background: graph spectral filtering}
\label{sec:spectral_filter}

We assume a graph with $M$ nodes to be characterized by a symmetric adjacency matrix $\A \in \mathbb{R}^{M \times M}$ and refer to \textit{graph signal} $\X \in \mathbb{R}^{M \times F}$ as the instance of all features (vectors in $\mathbb{R}^F$) associated with the graph nodes.
Let $\L = \I_M - \D^{-1/2}\A\D^{-1/2}$ be the symmetrically normalized Laplacian of the graph (where $\D$ is the degree matrix), with spectral decomposition $\L = \sum_{m=1}^M \lambda_m \mathbf{u}_m \mathbf{u}^T_m$.
A graph filter is an operator that modifies the components of $\X$ on the eigenvectors basis of $\L$, according to a frequency response $h$ acting on each eigenvalue $\lambda_m$. 
The filtered graph signal reads
\begin{equation}
    \label{eq:spctral_filter}
    \begin{aligned}
    \bar{\X} & =  \sum_{m=1}^M h(\lambda_m) \mathbf{u}_m \mathbf{u}^T_m \X =\\
             & = \U \, \text{diag}[h(\lambda_1), \dots, h(\lambda_M)] \, \U^T \X.
    \end{aligned}
\end{equation}
This formulation inspired the seminal work of Bruna \emph{et al.}~\cite{bruna2013spectral} that implemented spectral graph convolutions in a neural network. 
Their GNN learns end-to-end the parameters of a filter implemented as $h = \mathbf{B}\mathbf{c}$, where $\mathbf{B} \in \mathbb{R}^{M \times K}$ is a cubic B-spline basis and $\mathbf{c} \in \mathbb{R}^K$ is a vector of control parameters. 
Such filters are not localized, since the full projection on the eigenvectors yields paths of infinite length and the filter accounts for interactions of each node with the whole graph, rather than those limited to the node neighborhood.
Since this contrasts with the local design of classic convolutional filters, a follow-up work~\cite{henaff2015deep} introduced a parametrization of the spectral filters with smooth coefficients to achieve spatial localization.
However, the main issue with the spectral filtering in Eq.~\eqref{eq:spctral_filter} is computational complexity: not only the eigendecomposition of $\L$ is computationally expensive, but a double product with $\U$ is computed whenever the filter is applied.
Notably, $\U$ in Eq.~\eqref{eq:spctral_filter} is full even when $\L$ is sparse. 
Finally, since these spectral filters depend on a specific Laplacian spectrum, they cannot be transferred to graphs with another structure. 
For this reason, this spectral GNN cannot be used in downstream tasks such as graph classification or graph regression, where each datum is a graph with a different topology.

\subsection{GNNs based on polynomial filters and limitations}
\label{sec:poly_and_limitations}

The desired filter response $h(\lambda)$ can be approximated by a polynomial of order $K$,
\begin{equation}
    \label{eq:poly_response}
    h_\text{POLY}(\lambda) = \sum_{k=0}^K w_k \lambda^k,
\end{equation}
which performs a weighted moving average of the graph signal \cite{tremblay2018design}.
These filters overcome important limitations of the spectral formulation, as they avoid the eigendecomposition and their parameters are independent of the Laplacian spectrum~\cite{zhang2018end}.
Polynomial filters are localized in the node space, since the output at each node in the filtered signal is a linear combination of the nodes with their $K$-hop neighborhoods.

The order of the polynomial $K$ is assumed to be small and independent of the number $M$ of nodes in the graph.

To express polynomial filters in the node space, we first recall that the $k$-th power of any diagonalizable matrix, such as the Laplacian, can be computed by taking the power of its eigenvalues, \emph{i.e.}, $\L^k = \U \, \text{diag}[\lambda_1^k, \dots, \lambda_M^k] \, \U^T$.
It follows that the filtering operation becomes
\begin{equation}
\label{eq:poly_filtering}
\begin{aligned}
    \bar{\X} &= \left( w_0\I + w_1\L + w_2\L^2 + \dots + w_K \L^K \right) \X = \\
             &= \sum_{k=0}^K w_k \L^k \X.
\end{aligned}
\end{equation}

Eq.~\eqref{eq:poly_response} and \eqref{eq:poly_filtering} represent a generic polynomial filter.
Among the existing classes of polynomials, Chebyshev polynomials are often used in signal processing as they attenuate unwanted oscillations around the cut-off frequencies~\cite{shuman2011chebyshev}, which, in our case, are the eigenvalues of the Laplacian.
Fast localized GNN filters can approximate the desired filter response by means of the Chebyshev expansion $T_k(x) = 2xT_{k-1}(x) - T_{k-2}(x)$~\cite{defferrard2016convolutional}, resulting in convolutional layers that perform the filtering operation
\begin{equation}
    \label{eq:cheb_exp}
    \bar \X = \sigma \left( \sum \limits_{k=0}^{K-1} T_k(\tilde{\L})\X\mathbf{W}_k \right), 
\end{equation}
where $\tilde{\L} = 2\L/\lambda_\text{max} - \I_M$, $\sigma(\cdot)$ is a non-linear activation (\emph{e.g.}, a sigmoid or a ReLU function), and $\mathbf{W}_k \in \mathbb{R}^{F_\text{in} \times F_\text{out}}$ are the $k$ trainable weight matrices that map the node features from $\mathbb{R}^{F_\text{in}}$ to $\mathbb{R}^{F_\text{out}}$.

The output of a $k$-degree polynomial filter is a linear combination of the input within each vertex's $k$-hop neighborhood. 
Since the input beyond the $k$-hop neighborhood has no impact on the output of the filtering operation, to capture larger structures on the graph it is necessary to adopt high-degree polynomials. 
However, high-degree polynomials have poor interpolatory and extrapolatory performance since they overfit the known graph frequencies, \emph{i.e.}, the eigenvalues of the Laplacian.
This hampers the GNN's generalization capability as it becomes sensitive to noise and small changes in the graph topology.
Moreover, evaluating a polynomial with a high degree is computationally expensive both during training and inference~\cite{isufi2016autoregressive}.
Finally, since polynomials are very smooth, they cannot model filter responses with sharp changes.

A particular first-order polynomial filter has been proposed by \cite{kipf2016semi} for semi-supervised node classification.
In their GNN model, called Graph Convolutional Network (GCN), the convolutional layer is a simplified version of a Chebyshev filter, obtained from Eq.~\eqref{eq:cheb_exp} by considering $K=1$ and by setting $\mathbf{W} = \mathbf{W}_0 = -\mathbf{W}_1$
\begin{equation}
    \label{eq:GCN_layer}
    \bar \X= \sigma\left(\hat{\A}\X\mathbf{W}\right).
\end{equation}
Additionally, $\tilde{\L}$ is replaced by $\hat{\A} = \tilde{\D}^{-1/2}\tilde{\A}\tilde{\D}^{-1/2}$, with $\tilde{\A} = \A + \gamma \I_M$ (usually, $\gamma=1$).
The modified adjacency matrix $\hat{\A}$ contains self-loops that compensate for the removal of the term of order 0 in the polynomial, by ensuring that a node is part of its first-order neighborhood and that its features are preserved (to some extent) after convolution.
Higher-order neighborhoods can be reached by stacking multiple GCN layers.
On one hand, GCNs reduce overfitting and the heavy computational load of Chebyshev filters with high-order polynomials.
On the other hand, since each GCN layer performs a Laplacian smoothing, after few convolutions the node features becomes too smoothed over the graph~\cite{li2018deeper} and the initial node features are lost.


\section{Rational filters for graph signals}
\label{sec:rational_filters}

An ARMA filter can approximate well any desired filter response $h(\lambda)$ thanks to a rational design that, compared to polynomial filters, can model a larger variety of filter shapes~\cite{tremblay2018design}.
The filter response of an ARMA filter of order $K$, denoted in the following as ARMA\textsubscript{K}, reads
\begin{equation}
 \label{eq:arma_transfer}
 h_\text{ARMA\textsubscript{K}}(\lambda) = \frac{\sum_{k=0}^{K-1} p_k \lambda^k}{1 + \sum_{k=1}^K q_k \lambda^k},
\end{equation}
which translates to the following filtering relation in the node space
\begin{equation}
 \label{eq:arma_node_space}
 \bar \X = \left(\I + \sum_{k=1}^K q_k \L^k \right)^{-1} \left( \sum_{k=0}^{K-1} p_k \L^k \right) \X.
\end{equation}

Notice that by setting $q_k = 0$, for every $k$, one recovers a polynomial filter, which is considered as the MA term of the model. The inclusion of the additional AR term encoded by these coefficients makes the ARMA model robust to noise and allows to capture longer dynamics on the graph since $\bar{\x}$ depends, in turn, on several steps of propagation of the node features.
This is the key to capturing longer dependencies and more global structures on the graph, compared to a polynomial filter with the same degree.

The matrix inversion in Eq.~\eqref{eq:arma_node_space} is slow to compute and yields a dense matrix that prevents us from using sparse multiplications to implement the GNN.
In this paper, we follow a straightforward approach to avoid computing the inverse, which can be easily extended to a neural network implementation.
Specifically, we approximate the effect of an ARMA\textsubscript{1} filter by iterating, until convergence, the first-order recursion
\begin{equation}
\label{eq:recursive_arma}
    \bar{\X}^{(t+1)} = a \mathbf{M}\bar{\X}^{(t)} + b\X,
\end{equation}
where 
\begin{equation}
\label{eq:M}
    \mathbf{M} = \frac{1}{2}(\lambda_\text{max} - \lambda_\text{min}) \I - \L.
\end{equation}

The recursion in Eq.~\eqref{eq:recursive_arma} is adopted in graph signal processing to apply a low-pass filter on a graph signal~\cite{loukas2015distributed,isufi2016autoregressive}, but it is also equivalent to the recurrent update used in Label Propagation~\cite{zhou2004learning} and Personalized Page Rank~\cite{page1999pagerank} to propagate information on a graph by means of a random walk with a restart probability.

Following the derivation in \cite{isufi2016autoregressive}, we analyze the frequency response of an ARMA\textsubscript{1} filter from the convergence of Eq.~\eqref{eq:recursive_arma}: 
\begin{equation}
    \label{eq_recursive_arma_exp}
    \bar{\X} = \lim_{t \rightarrow \infty} \left[ (a\mathbf{M})^t \bar{\X}^{(0)} + b \sum \limits_{i=0}^t (a\mathbf{M})^i\X \right].
\end{equation}
The eigenvectors of $\M$ and $\L$ are the same, while the eigenvalues are related as follows: $\mu_m = (\lambda_\text{max} - \lambda_\text{min})/2 - \lambda_m$, where $\mu_m$ and $\lambda_m$ represent the $m$-th eigenvalue of $\M$ and $\L$, respectively. 
Since $\mu_m \in [-1,1]$, for $\lvert a \rvert < 1$ the first term of Eq.~\eqref{eq_recursive_arma_exp}, $(a\mathbf{M})^t$, goes to zero when $t \rightarrow \infty$, regardless of the initial point $\bar{\X}^{(0)}$.
The second term, $b \sum_{i=0}^t (a\mathbf{M})^i$, is a geometric series that converges to the matrix $b(\I - a\mathbf{M})^{-1}$, with eigenvalues $b/(1-a\mu_m)$.
It follows that the frequency response of the ARMA\textsubscript{1} filter is
\begin{equation}
    \label{eq:arma1_response}
    h_{\text{ARMA}_1}(\mu_m) = \frac{b}{1 - a\mu_m}. 
\end{equation}

By summing $K$ ARMA\textsubscript{1} filters, it is possible to recover the analytical form of the ARMA\textsubscript{K} filter in Eq.~\eqref{eq:arma_node_space}.
The resulting filtering operation is
\begin{equation}
    \label{eq:arma_filter}
    \bar{\X} = \sum \limits_{k=1}^K \sum \limits_{m=1}^M \frac{b_k}{1 - a_k\mu_m} \mathbf{u}_m \mathbf{u}_m^T \X,
\end{equation}
with 
\begin{equation}
    h_{\text{ARMA}_{K}}(\mu_m) = \sum_{k=1}^K \frac{b_k}{1 - a_k \mu_m}.
\end{equation}

Different orders ($\leq K$) of the numerator and denominator in Eq.~\eqref{eq:arma_transfer} are trivially obtained by setting some coefficients to 0. 
It follows that an ARMA filter generalizes a polynomial filter when all coefficients $q_k$ are set to zero.

\section{The ARMA neural network layer} 
\label{sec:arma}

\begin{figure*}[!ht]
	\centering
	\includegraphics[keepaspectratio,width=.8\textwidth]{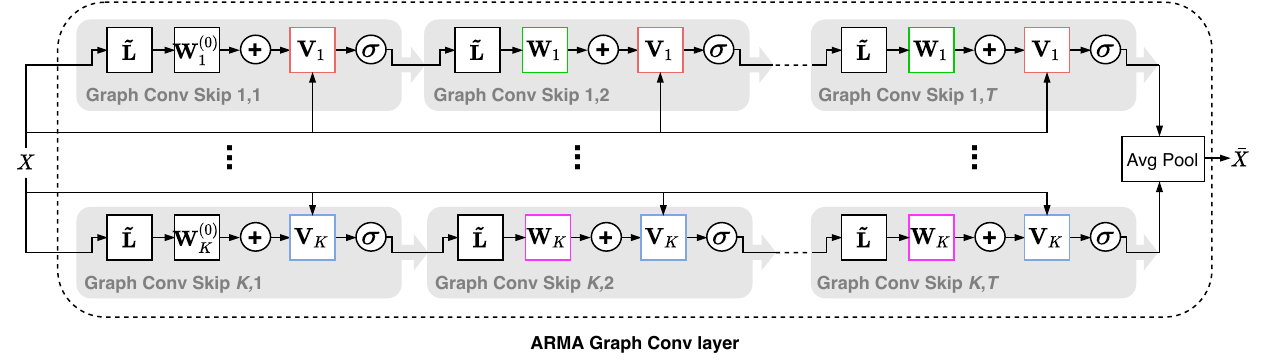}	
    \caption{The ARMA convolutional layer. Same color indicates that the weights are shared. }
	\label{fig:ARMA_block}
\end{figure*}

In graph signal processing, the filter coefficients $a$ and $b$ in Eq.~\eqref{eq:recursive_arma} are optimized with linear regression to reproduce a desired filter response $h^*(\lambda)$, which must be provided \textit{a priori} by the designer~\cite{isufi2016autoregressive}.
Here, we consider a machine learning approach that does not require to specify the target response $h^*(\lambda)$ but in which the parameters are learned end-to-end from the data by optimizing a task-dependent loss function.
Importantly, we also introduce non-linearities to enhance the representation capability of the filter response that can be learned. 

The proposed neural network formulation of the ARMA\textsubscript{1} filter implements the recursive update of Eq.~\eqref{eq:recursive_arma} with a \textit{Graph Convolutional Skip} (GCS) layer, defined as
\begin{equation}
\label{eq:graph_conv_skip}
    \bar{\X}^{(t+1)}= \sigma\left(\tilde{\L}\bar{\X}^{(t)}\mathbf{W} + \X\V\right),
\end{equation}
where $\sigma(\cdot)$ is a non-linearity such as ReLU, sigmoid, or hyperbolic tangent (\emph{tanh}), $\X$ are the initial node features, and $\W \in \mathbb{R}^{F_\text{out} \times F_\text{out}}$ and $\V \in \mathbb{R}^{F_\text{in} \times F_\text{out}}$ are trainable parameters.
The modified Laplacian matrix $\tilde{\L}$ is defined by setting $\lambda_\text{min}=0$ and $\lambda_\text{max}=2$ in Eq.~\eqref{eq:M} and then $\tilde{\L} = \mathbf{M}$. 
This is a reasonable simplification since the spectrum of $\L$ lies in $[0,2]$ and the trainable parameters $\W$ and $\V$ can compensate for the small offset introduced.
The unfolded recursion in Eq.~\eqref{eq:graph_conv_skip} corresponds to a stack of GCS layers that share the same parameters.

Each GCS layer is localized in the node space, as it performs a filtering operation that depends on local exchanges among neighboring nodes and, through the skip connection, also on the initial node features $\X$.
The computational complexity of the GCS layer is linear in the number of edges (both in time and space) since the layer can be efficiently implemented as a sparse multiplication between $\tilde \L$ and $\bar{\X}^{(t)}$.

The neural network formulation of an ARMA\textsubscript{1} filter is obtained by iterating Eq.~\eqref{eq:graph_conv_skip} until convergence, \emph{i.e.}, until $\| \bar{\X}^{(T+1)} - \bar{\X}^{(T)} \| < \epsilon$, where $\epsilon$ is a small positive constant and $T$ is the convergence time.
The convergence of the update in Eq.~\eqref{eq:graph_conv_skip}, which draws a connection to the original recursive formulation of the ARMA\textsubscript{1} filter, is guaranteed by Theorem \ref{th:1}.

\begin{theorem}
\label{th:1}
It is sufficient that $\| \W \|_2 < 1$ and that $\sigma(\cdot)$ is a non-expansive map for Eq.~\eqref{eq:graph_conv_skip} to converge to a unique fixed point, regardless of the initial state $\bar{\X}^{(0)}$.
\end{theorem}

\begin{proof}
Let $\bar \X_a^{(0)}$ and $\bar \X_b^{(0)}$ be two different initial states and $\left\Vert \W \right\Vert_2 < 1$.
After applying Eq.~\eqref{eq:graph_conv_skip} for $t + 1$ steps, we obtain states $\X_a^{(t + 1)}$ and $\bar \X_b^{(t + 1)}$.
If the non-linearity $\sigma(\cdot)$ is a non-expansive map, such as the ReLU function, the following inequality holds:

\begin{equation}
\label{eq:proof_1}
\begin{aligned}
& \left\Vert \bar \X_a^{(t + 1)} - \bar \X_b^{(t + 1)} \right\Vert_2 = \\
& = \left\Vert \sigma\left(\tilde{\L}\bar{\X}_a^{(t)}\mathbf{W} + \X\V \right) - \sigma\left(\tilde{\L}\bar{\X}_b^{(t)}\mathbf{W} + \X\V\right) \right\Vert_2 \leq \\
& \leq \left\Vert \tilde{\L}\bar{\X}_a^{(t)}\mathbf{W} + \X\V - \tilde{\L}\bar{\X}_b^{(t)}\mathbf{W} - \X\V \right\Vert_2 = \\
& = \left\Vert \tilde{\L}\bar{\X}_a^{(t)}\mathbf{W} - \tilde{\L}\bar{\X}_b^{(t)}\mathbf{W} \right\Vert_2 \leq\\
& \leq \left\Vert \tilde{\L} \right\Vert_2 \left\Vert \W \right\Vert_2 \left\Vert \bar{\X}_a^{(t)} - \bar{\X}_b^{(t)} \right\Vert_2.
\end{aligned}
\end{equation}

If the non-linearity $\sigma(\cdot)$ is also a squashing function (\emph{e.g.}, sigmoid or \textit{tanh}), then the first inequality in \eqref{eq:proof_1} is strict.

Since the largest singular value of $\tilde{\L}$ is $\leq 1$ by definition, it follows that $\left\Vert \tilde{\L} \right\Vert_2 \left\Vert \W \right\Vert_2 < 1$ and, therefore, \eqref{eq:proof_1} implies that Eq.~\eqref{eq:graph_conv_skip} is a contraction mapping.
The convergence to a unique fixed point and, thus, the inconsequentiality of the initial state, follow by the Banach fixed-point theorem~\cite{goebel1972fixed}.
\end{proof}

From Theorem \ref{th:1} it follows that it is possible to choose an arbitrary $\epsilon > 0$ for which
$$
    \exists T_\epsilon < \infty \text{ s.t. }\left\Vert \bar \X^{(t + 1)} - \bar \X^{(t)} \right\Vert_2 \leq \epsilon, \forall t \geq T_\epsilon.
$$
Therefore, we can easily implement a stopping criterion for the iteration, which is met in finite time.

Similar to the formulation of the ARMA filter in Eq.~\eqref{eq:arma_filter}, the output of the ARMA\textsubscript{K} convolutional layer is obtained by combining the outputs of $K$ parallel stacks of GCS layers.

\subsection{Implementation}
Each GCS stack $k$ may require a different and possibly high number of iterations $T_k$ to converge, depending on the value of the node features $\X$ and the weight matrices $\W_k$ and $\V_k$.
This makes the implementation of the neural network cumbersome, because the computational graph is dynamic and changes every time the weight matrices are updated with gradient descent during training.
Moreover, to train the parameters with backpropagation through time the neural network must be unfolded many times if $T_k$ is large, introducing a high computational cost and the vanishing gradient issue~\cite{bianchi2017recurrent}.

One solution is to follow the approach of Reservoir Computing, where the weight matrices $\W_k$ and $\V_k$ in each stack are randomly initialized and left untrained~\cite{lukovsevivcius2009reservoir, gallicchio2020fast}.
We notice that the random weights initialization guarantees that the $K$ GCS stacks implement different filtering operations. 
To compensate for the lack of training, high-dimensional features are exploited to generate rich latent representations that disentangle the factors of variations in the data~\cite{tino2020dynamical}.
However, randomized architectures with high-dimensional feature spaces are memory inefficient and computationally expensive at inference time.

A second approach, considered in this work, is to drop the requirement of convergence altogether and fix the number of iterations to a constant value $T$, so that $T_k = T$ in each GCS stack $k$.
In this way, we obtain a GNN that is easy to implement, fast to train and evaluate, and not affected by stability issues.
Notably, the constraint $\| \W \|_2 < 1$ of Theorem \ref{th:1} can be relaxed by adding to the loss function an L\textsubscript{2} weight decay regularization term.

Even by stacking a small number $T$ of GCS layers, we expect the GNN to learn a large variety of node representations thanks to the non-linearity and the trainable parameters~\cite{raghu2017expressive}.
As non-linearity we adopt the ReLU function that, compared to the squashing non-linearities, improves training efficiency by facilitating the gradient flow~\cite{goodfellow2016deep}.

Given the limited number of iterations, the initial state $\bar \X^{(0)}$ now influences the final representation $\bar \X^{(T)}$. 
A natural choice is to initialize the state with $\bar \X^{(0)} = \boldsymbol{0} \in \mathbb{R}^{M \times F_\text{out}}$ or with a linear transformation of the node features $\bar \X^{(0)} = \X \W^{(0)}$, where $\W^{(0)} \in \mathbb{R}^{F_\text{in} \times F_\text{out}}$ replaces $\W$ in the first layer of the stack.
We adopted the latter initialization so that the node features are propagated also by the first GCS layer.
We also note that it is possible to set $\W^{(0)} = \V$ to reduce the number of trainable parameters.

The output of the ARMA\textsubscript{K} convolutional layer is computed as
\begin{equation}
\label{eq:ARMA_layer}
    \bar{\X} = \frac{1}{K}\sum \limits_{k=1}^K \bar{\X}_k^{(T)},
\end{equation}
where $\bar{\X}_k^{(T)}$ is the output of the last GCS layer in the $k$-th stack.
Fig.~\ref{fig:ARMA_block} depicts a scheme of the proposed ARMA graph convolutional layer.

To encourage each GCS stack to learn a filtering operation with a response different from the other stacks, we apply stochastic dropout to the skip connections $\X\V_k$ in each GCS layer.
This leads to learning a heterogeneous set of features that, when combined to form the output of the ARMA\textsubscript{K} layer, yield powerful and expressive node representations.
We notice that the parameter sharing in each layer of the GCS stack endows the GNN with a strong regularization that helps to prevent overfitting and greatly reduces the model complexity, in terms of the number of trainable parameters.
Finally, since the GCS stacks are independent of each other, the computation of an ARMA layer can be distributed across multiple processing units.

\subsection{Properties and relationship with other approaches}
Contrarily to filters defined directly in the spectral domain~\cite{bruna2013spectral}, ARMA filters do not explicitly depend on the eigenvectors and the eigenvalues of $\L$, making them robust to perturbations in the underlying graph structure. 
For this reason, as formally proven for generic rational filters~\cite{transfer2019}, the proposed ARMA filters are transferable, \emph{i.e.}, they can be applied to graphs with different topologies not seen during training.


The skip connections in our architecture allow stacking many GCS layers without the risk of over-smoothing the node features.
Due to the weight sharing, the ARMA architecture has similarities with the recurrent neural networks with residual connections used to process sequential data~\cite{wu2016google}.

Similarly to GNNs operating directly in the node domain~\cite{scarselli2009graph,gallicchio2010graph}, each GCS layer computes the filtered signal $\bar{\mathbf{x}}_i^{(t+1)}$ at vertex $i$ as a combination of signals $\mathbf{x}_j^{(t)}$ in its 1-hop neighborhood, $j \in \mathcal{N}(i)$. 
Such a commutative aggregation solves the problem of undefined vertex ordering and varying neighborhood sizes, making the proposed operator permutation equivariant.

The skip connections in ARMA inject in each GCS layer $t$ of the stack the initial node features $\X$. 
This is different from a skip connection that either takes the output of the previous layer $\X^{(t-1)}$ as input~\cite{pham2017column,hamilton2017inductive}, or connects all the layers in a GNN stack directly to the output~\cite{xu2018representation}.

The ARMA layer can naturally deal with a time-varying topology and graph signals~\cite{holme2015modern,grattarola2018learning} by replacing the constant term $\X$ in Eq.~\eqref{eq:graph_conv_skip} with a time-dependent input $\X^{(t)}$.

Finally, we discuss the relationship between the proposed ARMA GNN and CayleyNets~\cite{levie2017cayleynets}, a GNN architecture that also approximates the effect of a rational filter.
Specifically, the filtering operation of a Cayley polynomial in the node space is
\begin{equation}
    \label{eq:cayley}
    \bar \X = w_0\X + 2\text{Re} \left\{ \sum_{k=1}^{K} w_k  (\L + i\I)^k(\L - i\I)^{-k}  \right\} \X.
\end{equation}

To approximate the matrix inversion in Eq.~\eqref{eq:cayley} with a sequence of differentiable operations, CayleyNets adopt a fixed number $T$ of Jacobi iterations.
In practice, the Jacobi iterations approximate each term $(\L + i\I)(\L - i\I)^{-1}$ as a polynomial of order $T$ with fixed coefficients.
Therefore, the resulting filtering operation performed by a CayleyNet assumes the form
\begin{equation}
    \label{eq:cayleynet}
    \bar \X \approx \sigma \left( w_0\X +   2\text{Re} \left\{\sum_{k=1}^{K} w_k \left( \sum_{t=1}^{T} \hat{\L}^t \right)^k \right\} \X \right) ,
\end{equation}
where $\hat{\L}$ is an operator with the same sparsity pattern of $\L$.
We note that Eq.~\eqref{eq:cayley} and \eqref{eq:cayleynet} slightly simplify the original formulation presented by Levie \emph{et al.}~\cite{levie2017cayleynets}, but allow us to better understand what type of operation is actually performed by the CayleyNet.
Specifically, Eq.~\eqref{eq:cayleynet} implements a polynomial filter of order $KT$, such as the one in Eq.~\eqref{eq:poly_filtering}.

For this reason, CayleyNets share strong similarities with the Chebyshev filter in Eq.~\eqref{eq:cheb_exp}, as it uses a (high-order) polynomial to propagate the node features on the graph for $KT$ hops before applying the non-linearity.
On the other hand, each of the $K$ parallel stacks in the proposed ARMA layer propagates the current node representations $\bar{\X}^{(t)}$ only for 1 hop and combines them with the node features $\X$ before applying the non-linearity.

\section{Spectral analysis of the ARMA layer}

\begin{figure*}[!ht]
	\centering
	\subfigure[$\tilde h$ in the 1\textsuperscript{st} GCS stack]{
        \includegraphics[width=0.31\textwidth]{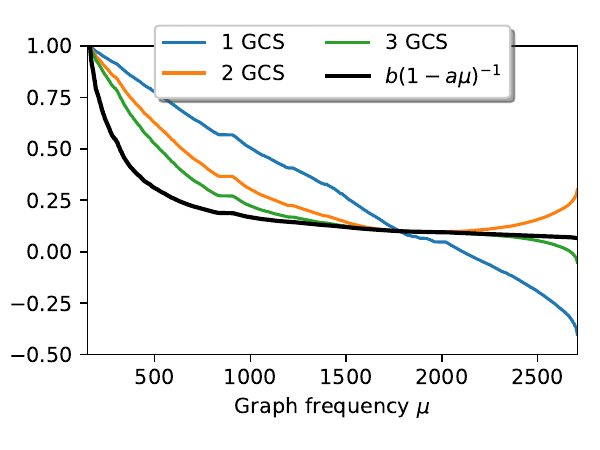}
    }
	~
	\subfigure[$\tilde h$ in the 2\textsuperscript{nd} GCS stack]{
        \includegraphics[width=0.31\textwidth]{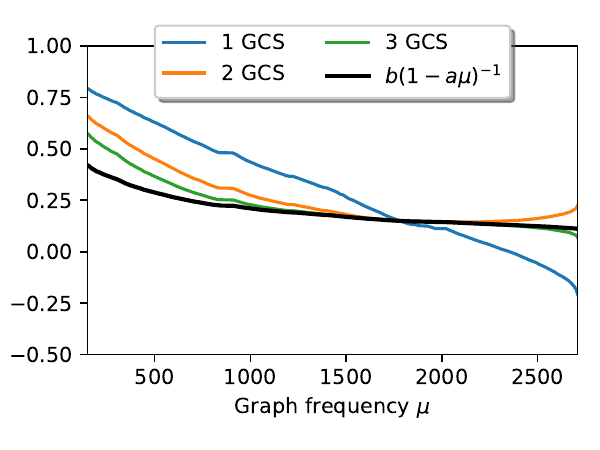}
    }
	~
	\subfigure[$\tilde h$ in a GCN stack]{
        \includegraphics[width=0.31\textwidth]{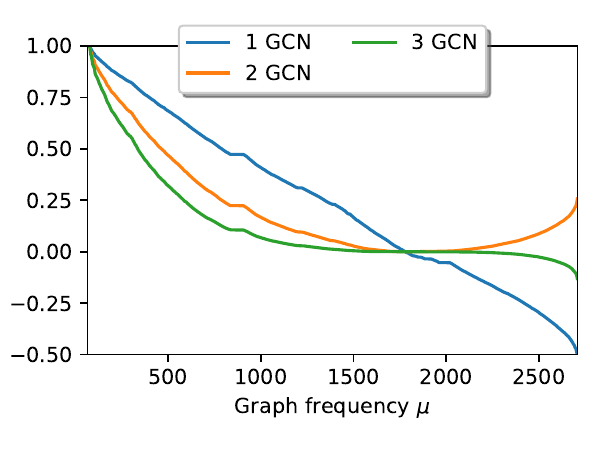}
    }

    \subfigure[Comp. of $\X$ and $\bar \X$ in the 1\textsuperscript{st} stack]{
        \includegraphics[width=0.31\textwidth]{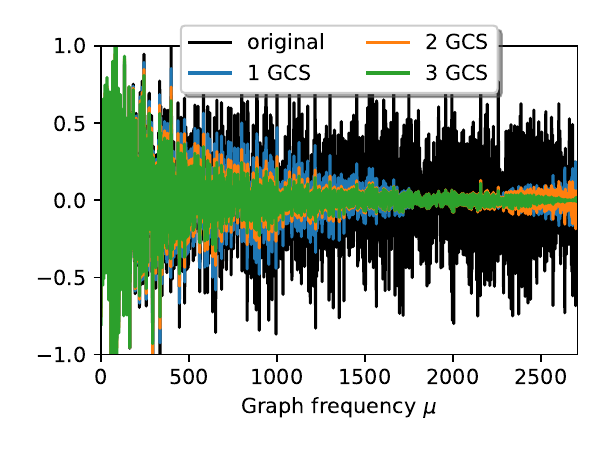}
    }
	~
	\subfigure[Comp. of $\X$ and $\bar \X$ in the 2\textsuperscript{nd} stack]{
        \includegraphics[width=0.31\textwidth]{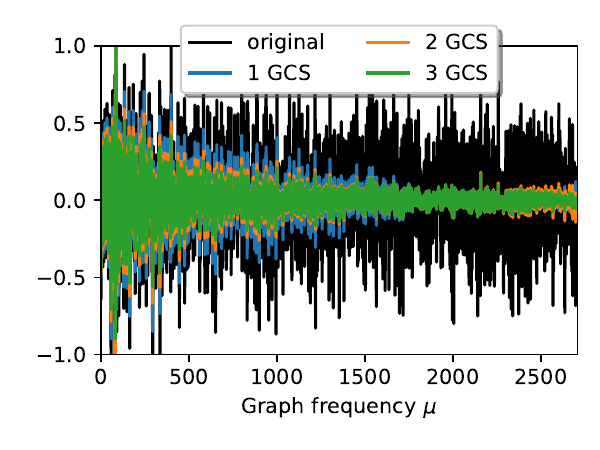}
    }
	~
	\subfigure[Comp. of $\X$ and $\bar \X$ in a GCN stack]{
        \includegraphics[width=0.31\textwidth]{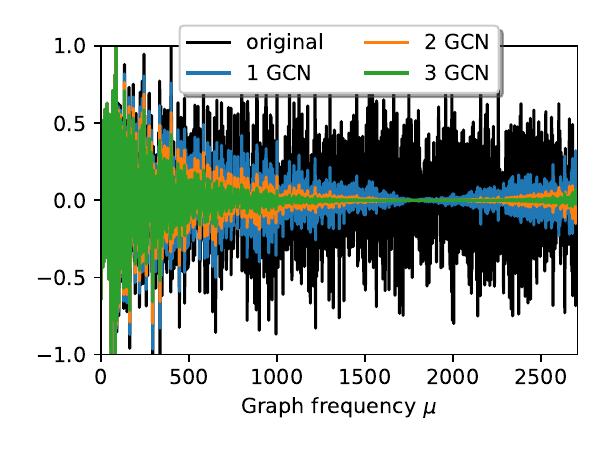}
    }

    \caption{\footnotesize In (a, b), the empirical filter responses of two GCS stacks for $T=1,2,3$; the black lines indicate the analytical response of an ARMA\textsubscript{1} filter with similar parameters. 
    In (c), the empirical response of a GCN with $T=1,2,3$ layers. 
    In (d, e), the original components of the input graph signal $\X$ (in black), and the components of the graph signal $\bar \X$ processed by two GCS stacks for $T=1,2,3$ (in color).
    In (f), the components of $\bar \X$ processed by a GCN with $T=1,2,3$ layers.}
	\label{fig:filters}
\end{figure*}

In this section we show how the proposed ARMA layer can implement filtering operations with a large variety of frequency responses.
The filter response of the ARMA filter derived in Sec.~\ref{sec:rational_filters} cannot be exploited to analyze our GNN formulation, due to the presence of non-linearities.
Therefore, we first recall that a filter changes the components of a graph signal $\X$ on the eigenbase induced by $\L$ (which is the same as the one induced by $\tilde{\L}$, according to Sylvester's theorem).
By referring to Eq.~\eqref{eq:spctral_filter}, $\X$ is first projected on the eigenspace of $\L$ by $\U^T$, then the filter $h(\lambda_m)$ changes the value of the component of $\X$ on each eigenvector $\u_m$, finally $\U^T$ maps back to the node space.
By left-multiplying $\U^T$ in Eq.~\eqref{eq:spctral_filter} we obtain
\begin{equation}
    \label{eq:spectral_filter_modified}
    \begin{aligned}
    \U^T \bar \X & = \text{diag}[h(\lambda_1), \dots, h(\lambda_M)] \, \U^T \X, \\
    \sum_{m=1}^M \mathbf{u}^T_m \bar \X & = \sum_{m=1}^M h(\lambda_m)  \mathbf{u}^T_m \X.
    \end{aligned}
\end{equation}

When $\bar \X$ is the output of the ARMA layer, the term $\U^T \bar \X$ defines how the original components, $\U^T \X$, are changed by the GNN.
Therefore, we can compute numerically the unknown filter response of the ARMA layer as the ratio between $\U^T \bar \X$ and $\U^T \X$.
We define the \textit{empirical filter response} $\tilde{h}$ as
\begin{equation}
    \label{eq:filter_approx}
    \begin{aligned}
    \tilde{h}_{m} &= \frac{F_\text{in}}{F_\text{out}} \frac{\sum_{f=1}^{F_\text{out}} \u_m^T \bar{\x}_{f}}{\sum_{f=1}^{F_\text{in}} \u_m^T\x_f}, \\
    \end{aligned}
\end{equation}

where $\bar{\x}_{f}$ is column $f$ of the output $\bar \X_k$, $\x_f$ is column $f$ of the graph signal $\X$, and $\u_m$ is an eigenvector of $\L$.

The empirical filter response allows us to analyze the type of filtering implemented by an ARMA layer.
We start by comparing the recursion in Eq.~\eqref{eq:recursive_arma}, which converges to an ARMA\textsubscript{1} filter with response $\{ h_{\text{ARMA}_1}(\mu_m) \}_{m=1}^M$ according to Eq.~\eqref{eq:arma1_response}, with the empirical response $\{ \tilde{h}_{m,k} \}_{m=1}^M$ of the $k$-th GCS stack.
To facilitate the interpretation of the results, we set the number of output features of the GCS layer to $F_\text{out} = 1$ by letting $\W = a$ and $\V = b\boldsymbol{1}_{F_\text{in}}$ in Eq.~\eqref{eq:graph_conv_skip}.
Notice that we are keeping the notation consistent with Eq.~\eqref{eq:recursive_arma}, where $a$ and $b$ are the parameters of the ARMA\textsubscript{1} filter.
In the following we consider the graph and the node features from the Cora citation network.
We remark that the examples in this section are not related to the results on the semi-supervised node classification task presented in Sec.~\ref{sec:experiments} and any other dataset could have been used instead of Cora.

Fig.~\ref{fig:filters}(a, b) show the empirical responses $\tilde{h}_1$ and $\tilde{h}_2$ of two different GCS stacks, when varying the number of layers $T$.
As $T$ increases, $\tilde{h}_1$ and $\tilde{h}_2$ become more similar to the analytical responses of the ARMA\textsubscript{1} filters, depicted as a black line in the two figures.
This supports our claim that $\tilde{h}$ can estimate the unknown response of the GNN filtering operation.

Fig.~\ref{fig:filters}(d, e) show how the two GCS stacks modify the components of $\X$ on the Fourier basis.
In particular, we depict in black the components $\u_m^T\X$, $m=1, \dots, M$ associated with each graph frequency $\mu_m$.
In colors, we depict the components $\u_m^T \bar\X$, which show how much the GCS stacks filter the components associated with each frequency.
The responses and the signal components in Fig.~\ref{fig:filters}(a) and \ref{fig:filters}(d) are obtained for $a=0.99$ and $b=0.1$, while in Fig.~\ref{fig:filters}(b) and \ref{fig:filters}(e) for $a=0.7$ and $b=0.15$.

In Fig~\ref{fig:filters}(c), we show the empirical response resulting from a stack of GCNs.
As also highlighted in recent work~\cite{wu2019simplifying, maehara2019revisiting}, the filtering obtained by stacking one or more GCNs has the undesired effect of symmetrically amplifying the lowest and also the highest frequencies of the spectrum. 
This is due to the GCN filter response, which is $(1-\lambda)^T$ in the linear case and can assume negative values when $T$ is odd.
The effect is mitigated by summing $\gamma \I_M$ to the adjacency matrix, which adds self-loops with weight $\gamma$ and shrinks the spectral domain of the graph filter.
For high values of $\gamma$, the GCN acts more as a low-pass filter that prevents high-frequency oscillations.
This is due to the self-loops that limit the spread of information across the graph and the communication between neighbors.
However, even after adding $\gamma \I_M$, GCN cuts almost completely the medium frequencies and then amplifies again the higher ones, as shown in Fig.~\ref{fig:filters}(f).

\begin{figure}[!t]
	\centering
	\subfigure[High-pass filter]{
        \includegraphics[width=0.46\columnwidth]{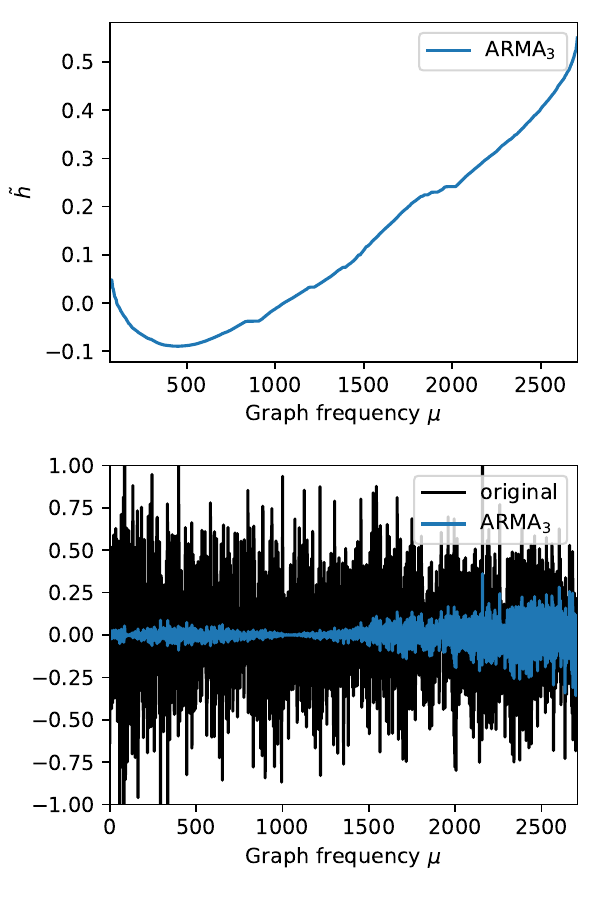}
    }
	~
	\subfigure[Band-pass filter]{
        \includegraphics[width=0.46\columnwidth]{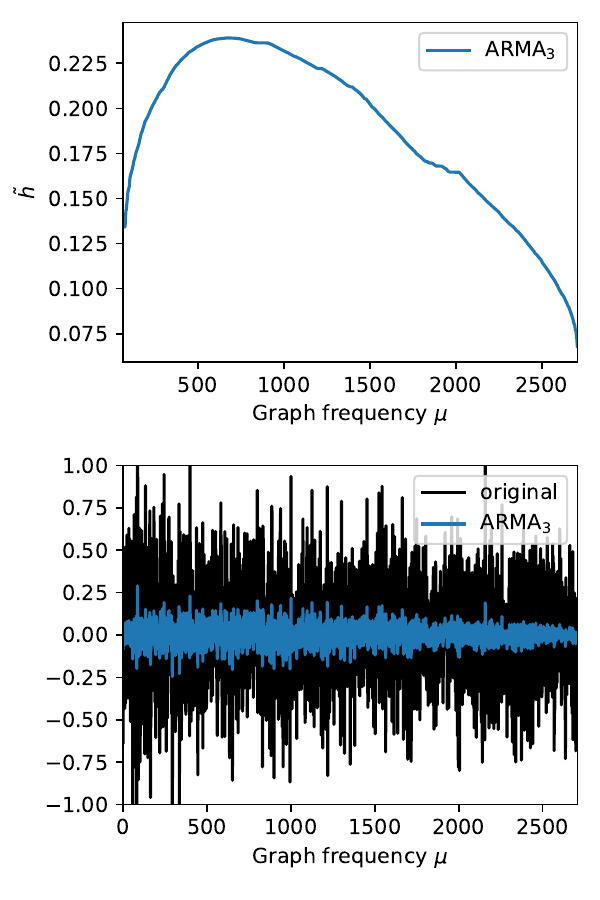}
    }
	
    \caption{\footnotesize While each GCS stack behaves as a low-pass filter, an ARMA layer with $K=3$ can implement filters of different shapes. 
    The ARMA layer in (a) implements a high-pass filtering operation that dampens low frequencies. 
    The ARMA layer in (b) implements a band-pass filtering operation that mostly allows medium frequencies.}
	\label{fig:filters2}
\end{figure}

A stack of GCNs lacks flexibility in implementing different filtering operations, as the only degree of freedom to modify a GCN's response consists of \textit{manually} tuning the hyperparameter $\gamma$ to shrink the spectrum.
On the other hand, different GCS stacks can generate heterogeneous filter responses, depending on the value of the trainable parameters in each stack.
This is what provides powerful modeling capability to the proposed ARMA layer, which can learn a large variety of filter responses that selectively shrink or amplify the Fourier components of the graph by combining $K$ GCS stacks. 

Similarly to an ARMA\textsubscript{1} filter, each GCS stack behaves as a low-pass filter that gradually dampens the Fourier components as their frequency increases.
However, we recall that high-pass and band-pass filters can be obtained as a linear combination of low-pass filters~\cite{oppenheim2001discrete}.
To show this behavior in practice, in Fig.~\ref{fig:filters2} we report the empirical filter responses and modified Fourier components obtained with two different ARMA\textsubscript{K} filters, for $K=3$.

\section{Experiments}
\label{sec:experiments}

We consider four downstream tasks: node classification, graph signal classification, graph classification, and graph regression.
Our experiments focus on comparing the proposed ARMA layer with GNNs layers based on polynomial filters, namely Chebyshev \cite{defferrard2016convolutional} and GCN \cite{kipf2016semi}, and CayleyNets~\cite{levie2017cayleynets} that, like ARMA, are based on rational spectral filters.
As additional baselines, we also include Graph Attention Networks (GAT)~\cite{velickovic2017graph}, GraphSAGE~\cite{hamilton2017inductive}, and Graph Isomorphism Networks (GIN)~\cite{xu2018powerful}.
The comparison with these methods helps to frame the proposed ARMA GNN within the current state of the art.
We also mention that other GNNs with graph convolutional filters related to our method have appeared while our work was under review~\cite{Ioannidis2020Pruned, NIPS2019_9016, zou2020graph, gao2019geometric}.

To ensure a fair and meaningful evaluation, we compare the performance obtained with a fixed GNN architecture, where we only change only the graph convolutional layers.
In particular, we fixed the GNN capacity (number of hidden units), used the same splits in each dataset, and the same training and evaluation procedures.
Finally, in all experiments we used the same polynomial order $K$ for polynomial/rational filters, or a stack of $K$ layers for GCN, GAT, GIN, and GraphSAGE layers.
The details of every dataset considered in the experiments and the optimal hyperparameters for each model are deferred to Sec.~\ref{sec:details}.

Public implementations of the ARMA layer are available in the open-source GNN libraries Spektral~\cite{grattarola2020graph} (TensorFlow/Keras) and PyTorch Geometric~\cite{fey2019fast} (PyTorch).

\subsection{Node classification}
First, we consider transductive node classification on three citation networks: Cora, Citeseer, and Pubmed. 
The input is a single graph described by an adjacency matrix $\mathbf{A} \in \mathbb{R}^{M \times M}$, the node features $\mathbf{X} \in \mathbb{R}^{M \times F_\text{in}}$, and the labels $\mathbf{y}_l \in \mathbb{R}^{M_l}$ of a subset of nodes $M_l \subset M$. 
The targets are the labels $\mathbf{y}_u \in \mathbb{R}^{M_u}$ of the unlabelled nodes.
The node features are sparse bag-of-words vectors representing text documents. 
The binary undirected edges in $\mathbf{A}$ indicate citation links between documents.
The models are trained using 20 labels per document class ($\mathbf{y}_l$) and the performance is evaluated as classification accuracy on $\mathbf{y}_u$.

Secondly, we perform inductive node classification on the protein-protein interaction (PPI) network dataset.
The dataset consists of 20 graphs used for training, 2 for validation, and 2 for testing. 
Contrarily to the transductive setting, the testing graphs (and the associated node features) are not observed during training. 
Additionally, each node can belong to more than one class (multi-label classification).

We use a 2-layers GNN with 16 hidden units for the citation networks and 64 units for PPI. 
In the citation networks high dropout rates and L\textsubscript{2}-norm regularization are exploited to prevent overfitting. 
Tab.~\ref{tab:nodecl_res} reports the classification accuracy obtained by a GNN equipped with different graph convolutional layers.

Transductive node classification is a semi-supervised task that demands using a simple model with strong regularization to avoid overfitting on the few labels available.
This is the key of GCN's success when compared to more complex filters, such as Chebyshev.
Thanks to its flexible formulation, the proposed ARMA layer can implement the right degree of complexity and performs well on each task. 
On the other hand, since the PPI dataset is larger and more labels are available during training, less regularization is required and the more complex models are advantaged. 
This is reflected by the better performance achieved by Chebyshev filters and CayleyNets, compared to GCN.
On PPI, ARMA significantly outperforms every other model, due to its powerful modeling capability that allows learning filter responses with different shapes.
Since each layer in GAT, GraphSAGE, and GIN combines the features of a node only with those from its 1\textsuperscript{st} order neighborhood, similarly to a GCN, these architectures need to stack more layers to reach higher-order neighborhoods and suffer from the same oversmoothing issue.

We notice that the optimal depth $T$ of the ARMA layer reported in Tab.~\ref{tab:hyper_nc} is low in every dataset.
We argue that a reason is the small average shortest path in the graphs (see Tab.~\ref{tab:nc_dataset}).
Indeed, most nodes in the graphs can be reached with only a few propagation steps, which is not surprising since many real networks are small-world~\cite{watts1998collective}.

\begin{figure}[!ht]
	\centering
	\caption{Training times on the PPI dataset, obtained with an Nvidia Titan Xp GPU.}
	\includegraphics[keepaspectratio,width=0.6\columnwidth]{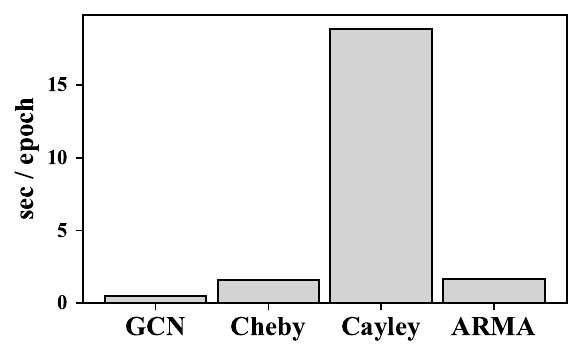}	
	\label{fig:train_time}
\end{figure}

Fig.~\ref{fig:train_time} shows the training times of the GNN model configured with GCN, Chebyshev, CayleyNet, and ARMA layers. 
The ARMA layer exploits sparse operations that are linear in the number of nodes in $\L$ and can be trained in a time comparable to a Chebyshev filter.
On the other hand, CayleyNet is slower than other methods, due to the complex formulation based on the Jacobi iterations that results in a high order polynomial. 

\begin{table}
\caption{Node classification accuracy.}
\setlength\tabcolsep{.6em} 
\small
\centering
\bgroup
\def\arraystretch{1.0} 
\begin{tabular}{lcccc}
\cmidrule[1.5pt]{1-5}
\textbf{Method} & \textbf{Cora} & \textbf{Citeseer} & \textbf{Pubmed} & \textbf{PPI} \\
\cmidrule[.5pt]{1-5}
GAT                     & 83.1 {\tiny$\pm$0.6} & 70.9 {\tiny$\pm$0.6} & 78.5 {\tiny$\pm$0.3} & 81.3 {\tiny$\pm$0.1} \\ 
GraphSAGE               & 73.7 {\tiny$\pm$1.8} & 65.9 {\tiny$\pm$0.9} & 78.5 {\tiny$\pm$0.6} & 70.0 {\tiny$\pm$0.0} \\ 
GIN                     & 75.1 {\tiny$\pm$1.7} & 63.1 {\tiny$\pm$2.0} & 77.1 {\tiny$\pm$0.7} & 78.1 {\tiny$\pm$2.6} \\ 
\cmidrule[.5pt]{1-5}
GCN                     & 81.5 {\tiny$\pm$0.4} & 70.1 {\tiny$\pm$0.7} & \textbf{79.0 {\tiny$\pm$0.5}} & 80.8 {\tiny$\pm$0.1} \\ 
Chebyshev               & 79.5 {\tiny$\pm$1.2} & 70.1 {\tiny$\pm$0.8} & 74.4 {\tiny$\pm$1.1} & 86.4 {\tiny$\pm$0.1} \\ 
CayleyNet               & 81.2 {\tiny$\pm$1.2} & 67.1 {\tiny$\pm$2.4} & 75.6 {\tiny$\pm$3.6} & 84.9 {\tiny$\pm$1.2} \\ 
\textbf{ARMA}           & \textbf{83.4 {\tiny$\pm$0.6}} & \textbf{72.5 {\tiny$\pm$0.4}} & 78.9 {\tiny$\pm$0.3} & \textbf{90.5 {\tiny$\pm$0.3}} \\ 
\cmidrule[1.5pt]{1-5}
\end{tabular}
\label{tab:nodecl_res}
\egroup
\end{table}

\subsection{Graph signal classification}
In this task, $N$ different graph signals $\mathbf{X}_n \in \mathbb{R}^{M \times F_\text{in}}, n=1,\dots,N$, defined on the same graph with adjacency matrix $\mathbf{A} \in \mathbb{R}^{M \times M}$, must be mapped to labels $y_1, \dots, y_N$.
We perform these experiments following the same setting of \cite{defferrard2016convolutional} for the MNIST and 20news datasets.

\textbf{MNIST.} To emulate a classic CNNs operating on a regular 2D grid, an 8-NN graph is defined on the 784 pixels of the MNIST images. 
To determine if a vertex $v_j$ belongs to the neighborhood $\mathcal{N}(v_i)$ of a vertex $v_i$, we compute the Euclidean distance between the 2D coordinates of pixels (vertices) $i$ and $j$.
The elements in $\A$ are  
\begin{equation}
\label{eq:edges}
    a_{ij} = 
    \begin{cases}
    1 & \text{if} \;\; v_j \in \mathcal{N}(v_i); \\
    0 & \text{otherwise}.
    \end{cases}
\end{equation}

Each graph signal is a vectorized image $\x \in \mathbb{R}^{784}$.
The architecture is a GNN(32)-P(4)-GNN(64)-P(4)-FC(512)-FC\textsubscript{Softmax}(10), where GNN($n$) indicates a GNN layer with $n$ filters, P($s$) a pooling operation with stride $s$, and FC($u$) a fully connected layer with $u$ units (the last FC has a Softmax activation).
Pooling is implemented by a hierarchical spectral clustering algorithm (GRACLUS)~\cite{dhillon2007weighted}, which maps the graph signal $\bar{\x}^{(l)}$ at layer $l$ into a new node feature space $\x^{(l+1)} \in \mathbb{R}^{M_{l+1} \times F_{l+1}}$. 

\begin{table}
\caption{Graph signal classification accuracy.}
\setlength\tabcolsep{.7em} 
\small
\centering
\bgroup
\def\arraystretch{1.0} 
\begin{tabular}{lcc}
\cmidrule[1.5pt]{1-3}
\textbf{GNN layer} & \textbf{MNIST} & \textbf{20news} \\
\cmidrule[.5pt]{1-3}
GCN             & 98.48 {\tiny$\pm$ 0.2} & 65.45 {\tiny$\pm$ 0.2} \\
Chebyshev       & 99.14 {\tiny$\pm$ 0.1} & 68.24 {\tiny$\pm$ 0.2} \\
CayleyNet       & 99.18 {\tiny$\pm$ 0.1} & 68.84 {\tiny$\pm$ 0.3} \\
\textbf{ARMA}   & \textbf{99.20 {\tiny$\pm$ 0.1}} & \textbf{70.02 {\tiny$\pm$ 0.1}} \\
\cmidrule[1.5pt]{1-3}
\end{tabular}
\label{tab:gsc_res}
\egroup
\end{table}

Tab.~\ref{tab:gsc_res} reports the results obtained by using GCN, Chebyshev, CayleyNet, or ARMA.
The results are averaged over 10 runs and show that ARMA achieves a slightly higher, and almost perfect, accuracy compared to Chebyshev and CayleyNet, while the performance of GCN is significantly lower.
Similarly to the PPI experiment, the larger amount of data allows more powerful architectures to be trained more precisely and to achieve better performance compared to the simpler GCN.

\textbf{20news.}
The dataset consists of 18,846 documents divided into 20 classes.
Each graph signal is a document represented by a bag-of-words of the $10^4$ most frequent words in the corpus, embedded via Word2vec \cite{mikolov2013efficient}. 
The underlying graph of $10^4$ nodes is defined by a 16-NN adjacency matrix built as in Eq.~\eqref{eq:edges}, with the difference that the vertex neighborhoods are computed from the Euclidean distance between the embeddings vectors rather than the pixel coordinates.
We report results obtained with a single convolutional layer (GCN, Chebyshev, CayleyNet, or ARMA), followed by global average pooling and Softmax. 
As in \cite{defferrard2016convolutional}, we use 32 channels for Chebyshev. 
Instead, for GCN, CayleyNet, and ARMA, better results are obtained with only 16 filters.
The classification accuracy reported in Tab.~\ref{tab:gsc_res} shows that ARMA significantly outperforms every other model also on this dataset.

For this experiment we used a particular configuration of the ARMA layer with $K=1$ and $T=1$ (see Tab.~\ref{tab:hyper_gsc}), which is equivalent to a GCN with a skip connection.
The skip connection allows to weight differently the contribution of the original node feature, compared to the features of the neighbors.
It is important to notice that, contrary to other downstream tasks, the 20news graph is generated from the similarity of word embeddings. 
Such an artificial graph always links an embedding vector to its first 16 neighbors.
We argue that, for some words, the links might be not very relevant and using a skip connection allows weighting them less.

Similarly to the node classification datasets, the average shortest path in the 20news graph is low (see Tab.~\ref{tab:stats_gsc}).
On the other hand, the MNIST graph has a much higher diameter, due to its regular structure with very localized connectivity.
This could explain why the optimal depth $T$ of the ARMA layer is larger for MNIST than for any other task (see Tab.~\ref{tab:hyper_gsc}), as several steps are necessary to mix the node features on the graph.

\subsection{Graph classification}
In this task, the $i$-th datum is a graph represented by a pair $\{ \A_i, \X_i \}, i=1, \dots N$, where $\A_i \in \mathbb{R}^{M_i \times M_i}$ is an adjacency matrix with $M_i$ nodes, and $\X_i \in \mathbb{R}^{M_i \times F}$ are the node features.
Each sample must be classified with a label $y_i$.
We test the models on five different datasets.
We use node degree, clustering coefficients, and node labels as additional node features.
For each dataset we adopt a fixed network architecture GNN-GNN-GNN-AvgPool-FC\textsubscript{Softmax}, where AvgPool indicates a global average pooling layer.
We compute the model performance with nested 10-fold cross-validation repeated for 10 runs, using $10\%$ of the training set in each fold for early stopping. 
Tab.~\ref{tab:gc_res} reports the average accuracy and includes the results obtained also by using GAT, GraphSAGE, and GIN as convolutional layers.
The GNN equipped with the proposed ARMA layer achieves the highest mean accuracy compared the polynomial filters (Chebyshev and GCN).
Compared to CayleyNets, which are also based on a rational filter implementation, ARMA achieves not only a higher mean accuracy but also a lower standard deviation.
These empirical results indicate that our implementation is robust and confirm the transferability of the proposed ARMA layer, discussed in Sec.~\ref{sec:arma}.

\begin{table}
\caption{Graph classification accuracy.}
\setlength\tabcolsep{.2em} 
\small
\centering
\bgroup
\def\arraystretch{1.0} 
\begin{tabular}{lccccc}
\cmidrule[1.5pt]{1-6}
\textbf{Method} & \textbf{Enzymes} & \textbf{Proteins} & \textbf{D\&D} & \textbf{MUTAG} & \textbf{BHard} \\
\midrule
GAT             & 51.7\tiny{$\pm 4.3$}  & 72.3\tiny{$\pm 3.1$} & 70.9\tiny{$\pm 4.0$}  & 87.3\tiny{$\pm 5.3$}  & 30.1\tiny{$\pm0.7$} \\
GraphSAGE       & 60.3\tiny{$\pm 7.1$}  & 70.2\tiny{$\pm 3.9$} & 73.6\tiny{$\pm 4.1$}  & 85.7\tiny{$\pm 4.7$}  & 71.8\tiny{$\pm1.0$} \\
GIN             & 45.7\tiny{$\pm 7.7$}  & 71.4\tiny{$\pm 4.5$} & 71.2\tiny{$\pm 5.4$}  & 86.3\tiny{$\pm 9.1$}  & 72.1\tiny{$\pm1.1$} \\
\midrule
GCN             & 53.0\tiny{$\pm 5.3$}  & 71.0\tiny{$\pm 2.7$} & 74.7\tiny{$\pm 3.8$}  & 85.7\tiny{$\pm 6.6$}  & 71.9\tiny{$\pm1.2$} \\
Chebyshev       & 57.9\tiny{$\pm 2.6$}  & 72.1\tiny{$\pm 3.5$} & 73.7\tiny{$\pm 3.7$}  & 82.6\tiny{$\pm 5.2$}  & 71.3\tiny{$\pm1.2$} \\
CayleyNet       & 43.1\tiny{$\pm 10.7$} & 65.6\tiny{$\pm 5.7$} & 70.3\tiny{$\pm 11.6$} & 87.8\tiny{$\pm 10.0$} & 70.7\tiny{$\pm2.4$} \\
\textbf{ARMA}   & \textbf{60.6\tiny{$\pm 7.2$}}  & \textbf{73.7\tiny{$\pm 3.4$}} & \textbf{77.6\tiny{$\pm 2.7$}}  & \textbf{91.5\tiny{$\pm 4.2$}}  & \textbf{74.1\tiny{$\pm0.5$}} \\
\cmidrule[1.5pt]{1-6}
\end{tabular}
\egroup
\label{tab:gc_res}
\end{table}




\subsection{Graph regression}
This task is similar to graph classification, with the difference that the target output $y_i$ is now a real value, rather than a discrete class label.
We consider the QM9 chemical database~\cite{ramakrishnan2014quantum}, which contains more than 130,000 molecular graphs.
The nodes represent heavy atoms and the undirected edges the atomic bonds between them. 
Nodes have discrete attributes indicating one of four possible elements. 
The regression task consists of predicting a given chemical property of a molecule given its graph representation.
As for graph classification, we evaluate the performance on the 80-10-10 train-validation-test splits of the nested 10-folds.
The network architecture adopted to predict each property is GNN(64)-AvgPool-FC(128).
We report in Tab.~\ref{tab:gr_res} the mean squared error (MSE) averaged over 10 independent runs, relative to the prediction of 9 molecular properties.
It can be noticed that each model achieves a very low standard deviation. 
One reason is the very large amount of training data, which allows the GNN to learn a configuration that generalizes well.
Contrarily to the previous tasks, here there is not a clear winner among GCN, Chebyshev, and CayleyNet, since each of them performs better than the others on some tasks.
On the other hand, ARMA always achieves the lowest MSE in predicting each molecular property. 

\begin{table}
\caption{Graph regression mean squared error. }
\setlength\tabcolsep{.3em} 
\small
\centering
\bgroup
\def\arraystretch{1.0} 
\begin{tabular}{lcccc}
\cmidrule[1.5pt]{1-5}
\textbf{Property} & \textbf{GCN} & \textbf{Chebyshev} & \textbf{CayleyNet} & \textbf{ARMA} \\
\midrule
mu       & 0.445\tiny{$\pm0.007$} & 0.433\tiny{$\pm0.003$} & 0.442\tiny{$\pm0.009$} & \textbf{0.394\tiny{$\pm0.005$}} \\
alpha    & 0.141\tiny{$\pm0.016$} & 0.171\tiny{$\pm0.008$} & 0.118\tiny{$\pm0.005$} & \textbf{0.098\tiny{$\pm0.005$}} \\
HOMO     & 0.371\tiny{$\pm0.030$} & 0.391\tiny{$\pm0.012$} & 0.336\tiny{$\pm0.007$} & \textbf{0.326\tiny{$\pm0.010$}} \\
LUMO     & 0.584\tiny{$\pm0.051$} & 0.528\tiny{$\pm0.005$} & 0.679\tiny{$\pm0.148$} & \textbf{0.508\tiny{$\pm0.011$}} \\
gap      & 0.650\tiny{$\pm0.070$} & 0.565\tiny{$\pm0.015$} & 0.758\tiny{$\pm0.106$} & \textbf{0.552\tiny{$\pm0.013$}} \\
R2       & 0.132\tiny{$\pm0.005$} & 0.294\tiny{$\pm0.022$} & 0.185\tiny{$\pm0.043$} & \textbf{0.119\tiny{$\pm0.019$}} \\
ZPVE     & 0.349\tiny{$\pm0.022$} & 0.358\tiny{$\pm0.001$} & 0.555\tiny{$\pm0.174$} & \textbf{0.338\tiny{$\pm0.001$}} \\
U0\_atom & 0.064\tiny{$\pm0.003$} & 0.126\tiny{$\pm0.017$} & 1.493\tiny{$\pm1.414$} & \textbf{0.053\tiny{$\pm0.004$}} \\
Cv       & 0.192\tiny{$\pm0.012$} & 0.215\tiny{$\pm0.010$} & 0.184\tiny{$\pm0.009$} & \textbf{0.163\tiny{$\pm0.007$}} \\
\cmidrule[1.5pt]{1-5}
\end{tabular}
\egroup
\label{tab:gr_res}
\end{table}


\section{Experimental details}
\label{sec:details}

\subsection{Node classification}

Tab.~\ref{tab:nc_dataset} reports for each node classification dataset the number of nodes, number of edges, number of node attributes (size of the node feature vectors), average shortest path of the graph (Avg. SP), and the number of classes that each node can be assigned to.
The three citation networks (Cora, Citeseer, and Pubmed) are taken from \url{https://github.com/tkipf/gcn/raw/master/gcn/data/}, while the PPI dataset is taken from \url{http://snap.stanford.edu/graphsage/}.

\bgroup
\def\arraystretch{1.0} 
\setlength\tabcolsep{.2em} 
\begin{table}[!ht]
\small
\centering
\caption{Summary of the node classification datasets.} 
\label{tab:nc_dataset}
\begin{tabular}{lccccc}
\cmidrule[1.5pt]{1-6}
\textbf{Dataset} & \textbf{Nodes} & \textbf{Edges} & \textbf{Node attr.} & \textbf{Avg. SP} & \textbf{Node classes} \\
\cmidrule[.5pt]{1-6}
Cora     & 2708  & 5429   & 1433 & 5.87\tiny{$\pm$1.52} & 7   (single label) \\
Citeseer & 3327  & 9228   & 3703 & 6.31\tiny{$\pm$2.00} & 6   (single label) \\
Pubmed   & 19717 & 88651  & 500  & 6.34\tiny{$\pm$1.22} & 3   (single label) \\
PPI      & 56944 & 818716 & 50   & 2.76\tiny{$\pm$0.56} & 121 (multi-label)  \\
\cmidrule[1.5pt]{1-6}
\end{tabular}
\end{table}
\egroup

\begin{table}[!ht]
\setlength\tabcolsep{.25em} 
\small
\centering
\caption{Hyperparameters for node classification.}
\bgroup
\def\arraystretch{1.0} 
\begin{tabular}{lccc|c|c|cc|cc}
\cmidrule[1.5pt]{1-10}
\multirow{ 2}{*}{\textbf{Dataset}} & \multirow{ 2}{*}{L\textsubscript{2} reg.} & \multirow{ 2}{*}{$p_\text{drop}$} & \multirow{ 2}{*}{lr} & \textbf{GCN} & \textbf{Cheby.} & \multicolumn{2}{c|}{\textbf{Cayley}} & \multicolumn{2}{c}{\textbf{ARMA}} \\
& & & & $L$ & $K$ & $K$ & $T$ & $K$ & $T$ \\
\midrule
Cora     & 5e-4 & 0.75 & 0.01 & 1 & 2 & 1 & 5 & 2 & 1  \\
Citeseer & 5e-4 & 0.75 & 0.01 & 1 & 3 & 1 & 5 & 3 & 1  \\
Pubmed   & 5e-4 & 0.25 & 0.01 & 1 & 3 & 2 & 5 & 1 & 1  \\
PPI      & 0.0  & 0.25 & 0.01 & 2 & 3 & 3 & 5 & 3 & 2  \\
\cmidrule[1.5pt]{1-10}
\end{tabular}
\label{tab:hyper_nc}
\egroup
\end{table}

Tab.~\ref{tab:hyper_nc} describes the optimal hyperparameters used in GCN, Chebyshev, CayleyNet, and ARMA for each node classification dataset. 
For all GNN, we report the L\textsubscript{2} regularization weight, the learning rate (lr) and dropout probability ($p_\text{drop}$). For GCN, we report the number of stacked graph convolutions ($L$). For Chebyshev, we report the polynomial order ($K$). 
For CayleyNet, we report the polynomial order ($K$) and the number of Jacobi iterations ($T$).
For ARMA, we report the number of GCS stacks ($K$) and the stacks' depth ($T$).
Additionally, we configured the MLP in GIN with 2 hidden layers and trained the parameter $\epsilon$, while for GraphSAGE we used the \textit{max} aggregator, to differentiate more its behavior from GCN and GIN.
Finally, GAT is configured with 8 attention heads and the same number of layers $L$ as GCN.

Each model is trained for 2000 epochs with early stopping (based on the validation accuracy) at 50 epochs.
We used full-batch training, \emph{i.e.}, in each epoch the weights are updated one time, according to a single batch that includes all the training data. 

\subsection{Graph regression}

The QM9 dataset used for graph regression is available at \url{http://quantum-machine.org/datasets/}, and its statistics are reported in Tab.~\ref{tab:gr_dataset}.

The hyperparameters are reported in Tab.~\ref{tab:hyper_gr}. 
Only for this task, CayleyNets use only 3 Jacobi iterations, since with more iterations we experienced numerical errors and the loss quickly diverged.
All models are trained for 1000 epochs with early stopping at 50 epochs, using the Adam optimizer with learning rate $10^{-3}$. 
We used batch size 64 and no L\textsubscript{2} regularization.

\bgroup
\def\arraystretch{1.0} 
\setlength\tabcolsep{.8em} 
\begin{table}[!ht]
\small
\centering
\caption{Summary of the graph regression dataset.} 
\label{tab:gr_dataset}
\begin{tabular}{cccc}
\cmidrule[1.5pt]{1-4}
\textbf{Samples} & \textbf{Avg. nodes} & \textbf{Avg. edges} & \textbf{Node attr.} \\
\cmidrule[.5pt]{1-4}
133,885 & 8.79 & 27.61 & 1 \\
\cmidrule[1.5pt]{1-4}
\end{tabular}
\end{table}
\egroup

\begin{table}[!ht]
\setlength\tabcolsep{.5em} 
\small
\centering
\caption{Hyperparameters for graph classification and graph regression.}
\bgroup
\def\arraystretch{1.0} 
\begin{tabular}{l|c|c|cc|ccc}
\cmidrule[1.5pt]{1-8}
\multirow{ 2}{*}{\textbf{Dataset}} & \textbf{GCN} & \textbf{Cheby.} & \multicolumn{2}{c|}{\textbf{Cayley}} & \multicolumn{3}{c}{\textbf{ARMA}} \\
                                   & $L$ & $K$ & $K$ & $T$ & $p_{drop}$ & $K$ & $T$ \\
\midrule
QM9      & 3 & 3 & 3 & 3 & 0.75 & 3 & 3 \\
\cmidrule[1.5pt]{1-8}
\end{tabular}
\label{tab:hyper_gr}
\egroup
\end{table}

\subsection{Graph classification}

The datasets Enzymes, Proteins, D\&D, and MUTAG are taken from the Benchmark Data Sets for Graph Kernels \url{https://ls11-www.cs.tu-dortmund.de/staff/morris/graphkerneldatasets}, while the dataset Bench-hard is taken from \url{https://github.com/FilippoMB/Benchmark_dataset_for_graph_classification}.
The statistics of each graph classification dataset are summarized in Tab.~\ref{tab:gc_dataset}.

For all methods, we use a fixed architecture composed of three GNN layers, each with 32 output units, ReLU activations, and L\textsubscript{2} regularization with a factor of $10^{-4}$. 
All models are trained to convergence with Adam, using a learning rate of $10^{-3}$, batch size of 32, and patience of 50 epochs. 
We summarize in Tab.~\ref{tab:hyper_gc} the hyperparameters used for ARMA, Chebyshev, and CayleyNets on the different datasets. 

\bgroup
\def\arraystretch{1.0} 
\begin{table*}[!ht]
\small
\centering
\caption{Summary of the graph classification datasets.} 
\label{tab:gc_dataset}
\begin{tabular}{lcccccc}
\cmidrule[1.5pt]{1-7}
\textbf{Dataset} & \textbf{Samples} & \textbf{Classes} & \textbf{Avg. nodes} & \textbf{Avg. edges} & \textbf{Node attr.} & \textbf{Node labels} \\
\cmidrule[.5pt]{1-7}
Bench-hard   & 1,800  & 3  & 148.32 & 572.32 & -- & yes \\
Enzymes      & 600   & 6  & 32.63  & 62.14  & 18 & no  \\
Proteins     & 1,113  & 2  & 39.06  & 72.82  & 1  & no  \\
D\&D         & 1,178  & 2  & 284.32 & 715.66 & -- & yes \\
MUTAG        & 188   & 2  & 17.93  & 19.79  & -- & yes \\
\cmidrule[1.5pt]{1-7}
\end{tabular}
\end{table*}
\egroup

\begin{table}[!ht]
\setlength\tabcolsep{.5em} 
\small
\centering
\caption{Hyperparameters for graph classification and graph regression.}
\bgroup
\def\arraystretch{1.0} 
\begin{tabular}{l|c|c|cc|ccc}
\cmidrule[1.5pt]{1-8}
\multirow{2}{*}{\textbf{Dataset}} & \textbf{GCN} & \textbf{Cheby} & \multicolumn{2}{c|}{\textbf{Cayley}} & \multicolumn{3}{c}{\textbf{ARMA}} \\
                                   & $L$ & $K$ & $K$ & $T$ & $p_{drop}$ & $K$ & $T$            \\
\midrule
Bench-hard  & 2 & 2 & 2 & 10 & 0.4 & 1 & 2 \\
Enzymes     & 2 & 2 & 2 & 10 & 0.6 & 2 & 2 \\
Proteins    & 4 & 4 & 4 & 10 & 0.6 & 4 & 4 \\  
D\&D        & 4 & 4 & 4 & 10 & 0.0 & 4 & 4 \\
MUTAG       & 4 & 4 & 4 & 10 & 0.0 & 4 & 4 \\
\cmidrule[1.5pt]{1-8}
\end{tabular}
\label{tab:hyper_gc}
\egroup
\end{table}

\subsection{Graph signal classification}

To generate the datasets we used the code available at \url{github.com/mdeff/cnn_graph}.
The models are trained for 20 epochs on each dataset. We used batches of size 32 for MNIST and 128 for 20news.
In the 20news dataset, the word embeddedings have size 200.

\begin{table}[!ht]
\setlength\tabcolsep{.15em} 
\small
\centering
\caption{Summary of the graph signal classification datasets.}
\bgroup
\def\arraystretch{1.1} 
\begin{tabular}{lccccccc}
\cmidrule[1.5pt]{1-8}
\textbf{Dataset} & \textbf{Nodes} & \textbf{Edges} & \textbf{Avg. SP} & \textbf{Class} & \textbf{Train} & \textbf{Val} & \textbf{Test} \\
\cmidrule[.5pt]{1-8}
MNIST    & 784    & 5,928    & 12.36\tiny{$\pm$5.45} & 10 & 55$k$ & 5$k$ & 10$k$ \\
20news   & 10$k$ & 249,944  & 4.21\tiny{$\pm$0.94}  & 20 & 10,168 & 7,071 & 7,071  \\
\cmidrule[1.5pt]{1-8}
\end{tabular}
\label{tab:stats_gsc}
\egroup
\end{table}

\begin{table}[!ht]
\setlength\tabcolsep{.2em} 
\small
\centering
\caption{Hyperparameters for graph signal classification.}
\bgroup
\def\arraystretch{1.0} 
\begin{tabular}{lccc|c|c|cc|cc}
\cmidrule[1.5pt]{1-10}
\multirow{ 2}{*}{\textbf{Dataset}} & \multirow{ 2}{*}{L\textsubscript{2} reg.} & \multirow{ 2}{*}{lr} & \multirow{ 2}{*}{$p_\text{drop}$} & \textbf{GCN} & \textbf{Cheby.} & \multicolumn{2}{c|}{\textbf{Cayley}} & \multicolumn{2}{c}{\textbf{ARMA}} \\
& & & & $L$ & $K$ & $K$ & $T$ & $K$ & $T$ \\
\midrule
MNIST    & 5e-4 & 1e-3 & 0.5  & 3 & 25 & 12 & 11 & 5 & 10 \\ 
20news   & 1e-3 & 1e-3 & 0.7  & 1 & 5  & 5  & 10  & 1 & 1  \\ 
\cmidrule[1.5pt]{1-10}
\end{tabular}
\label{tab:hyper_gsc}
\egroup
\end{table}

Tab.~\ref{tab:stats_gsc} reports, for each graph signal classification dataset: the number of nodes and edges of the graph and the average shortest path (Avg. SP), the number of classes each graph signal can be assigned to, and the number of graph signals in the training, validation, and test set.
Tab.~\ref{tab:hyper_gsc} reports the optimal hyperparameters configuration for each model.

\section{Conclusions}
\label{sec:conclusion}

This paper introduced the ARMA layer, a novel graph convolutional layer based on a rational graph filter.
The ARMA layer models more expressive filter responses and can account for larger neighborhoods compared to GNN layers based on polynomial filters of the same order.
Our ARMA layer consists of parallel stacks of recurrent operations, which approximate a graph filter with an arbitrary order $K$ by means of efficient sparse tensor multiplications. 
We reported a spectral analysis of our neural network implementation, which provides valuable insights into the proposed method and shows that our ARMA layer can implement a large variety of filter responses.
The experiments showed that the proposed ARMA layer outperforms existing GNN architectures, including those based on polynomial filters and other more complex models, on a large variety of graph machine learning tasks. 

\bibliographystyle{icml2018}
\bibliography{references}

\end{document}